\documentclass[11pt,a4paper]{article}
\usepackage[hyperref]{eacl2021}
\usepackage{times}

\usepackage{microtype}

\usepackage{latexsym,graphicx,booktabs,multirow}
\usepackage{subfigure}
\usepackage{url,algorithm2e}

\definecolor{myorange}{RGB}{230, 159, 0}
\definecolor{myblue}{RGB}{0, 114,178}

\aclfinalcopy %
\def\aclpaperid{755} %

\title{We Need to Talk About Random Splits}

\author{Anders Søgaard$^{1,2}$ \, Sebastian Ebert$^1$ \, Jasmijn Bastings$^1$ \, Katja Filippova$^1$\\
  $^1$: Google Research, Berlin \\
  $^2$: University of Copenhagen \\
  $\{$\texttt{soegaard}$|${\tt eberts}$|${\tt bastings}$|${\tt katjaf}$\}${\tt @google.com} \\}

\date{}

\begin{document}
\maketitle
\begin{abstract}
\newcite{Gorman:Bedrick:19} argued for using random splits rather than standard splits in NLP experiments. We argue that random splits, like standard splits, lead to overly optimistic performance estimates. We can also split data in biased or adversarial ways, e.g., training on short sentences and evaluating on long ones. Biased  sampling has been used in domain adaptation to simulate real-world drift; this is known as the covariate shift assumption. In NLP, however, even worst-case splits, maximizing bias, often under-estimate the error observed on new samples of in-domain data, i.e., the data that models should minimally generalize to at test time. This invalidates the covariate shift assumption. Instead of using multiple random splits, future benchmarks should ideally include multiple, independent test sets instead; if infeasible, we argue that multiple biased splits leads to more realistic performance estimates than multiple random splits. %
 
\end{abstract}

\section{Introduction}
It is common practice in NLP to collect and annotate a text corpus -- and split it into training, development and test data.
These splits are often based on the order in which texts were published or sampled, and are referred to as `standard splits'. \newcite{Gorman:Bedrick:19} recently showed that system ranking results based on \emph{standard} splits differ from results based on \emph{random} splits and used this to argue in favor of using random splits. 
While perhaps less common, random splits are already used in probing \cite{Elazar:Goldberg:18}, interpretability \cite{Poerner:ea:18}, as well as core NLP tasks \cite{Yu:ea:19,Geva:ea:19}.\footnote{See also many of the tasks in the SemEval evaluation campaigns: {\url{http://alt.qcri.org/semeval2020/}}}

\begin{figure}[t]
\includegraphics[width=\columnwidth]{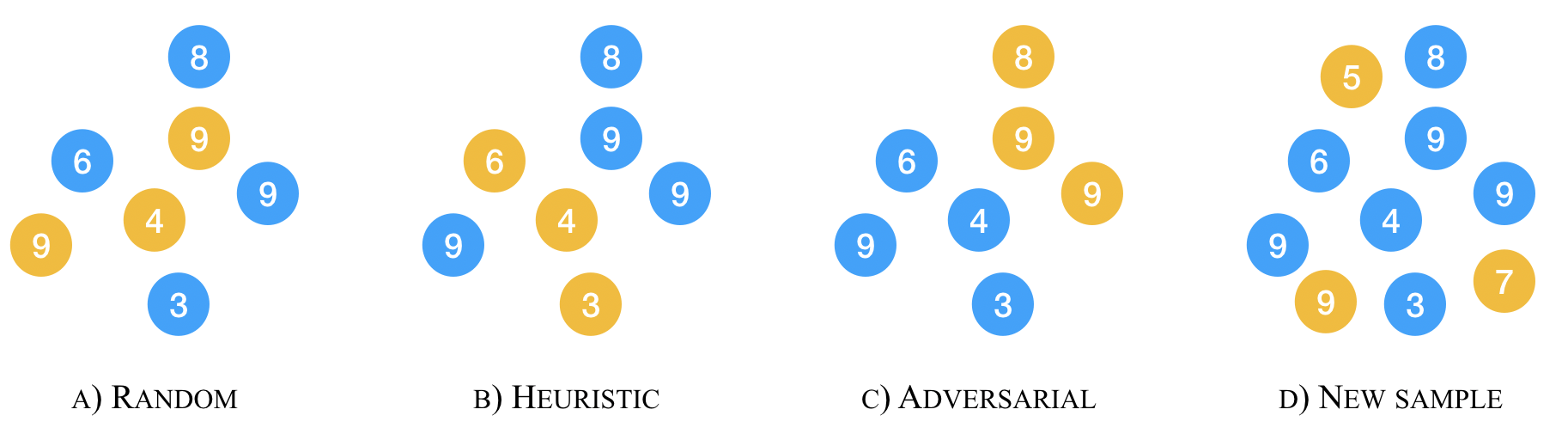}
     \caption{Data splitting strategies. Each ball  corresponds to a sentence represented in (two-dimensional) feature space. \textcolor{myblue}{Blue} (dark)/\textcolor{myorange}{orange} (bright) balls represent examples for \textcolor{myblue}{training}/\textcolor{myorange}{test}. Numbers represent sentence length. Heuristic splits can, e.g., be based on sentence length; adversarial splits maximize divergence.}
    \label{fig:splits}
\end{figure}

\newcite{Gorman:Bedrick:19} focus on whether there is a significant performance difference $\delta$ between systems $S_1$ and $S_2$; $\mathcal{M}(G_{\mathit{test}},S_1)-\mathcal{M}(G_{\mathit{test}},S_2)$, in their notation. They argue McNemar's test \cite{Gillick:Cox:89} or bootstrap \cite{Elfron:81} can establish that $\delta\neq 0$, using random splits to sample from $G_{\mathit{test}}$. This, of course, relies on the assumption that data is representative, i.e., was sampled i.i.d.~\cite{Wolpert:96}.

In reality, what \newcite{Gorman:Bedrick:19} call {\em the true difference in system performance}, i.e.,  $\delta=\mathcal{M}(G_{\mathit{test}},S_1)-\mathcal{M}(G_{\mathit{test}},S_2)$, is the system difference on {\em data that users would expect the systems to work well on} (see \S2~for practical examples) -- and not just on the corpus that we have annotations for. Our corpus-based estimates of $\delta$ can in fact be very misleading, i.e., very different from performance on new samples of data. In this paper, we investigate {\em how} misleading our estimates can be: We show that random splits consistently over-estimate performance at test time. This favors systems that overfit. We investigate alternatives across a heterogeneous set of NLP tasks. Based on our experiments, our answer to community-wide overfitting to standard splits is {\em not} to use random splits but to collect more diverse data with different biases -- or if that is not feasible, split your data in adversarial, not random, ways. In general, we observe that estimates of test time error are worst for random splits, slightly better for standard splits (if those exist), better for heuristic and adversarial splits, but error still tends to be higher on new (in-domain) samples; see Figure~\ref{fig:splits}. 

Our results not only refute the hypothesis that $\delta$ can be estimated using random splits \cite{Gorman:Bedrick:19},\footnote{Or cross-validation, as more recently proposed in \newcite{szymanski-gorman-2020-best}. In this very interesting follow-up paper, about Bayesian inference of $\delta$, the authors write that their 
"estimates are valid insofar as the data sets used to
estimate the Bayesian models comprise a representative sample of a coherent population of data sets." Our results show how {\em off} this assumption is.} but also the covariate shift hypothesis  \cite{Shimodaira:00,Shah:ea:20} that $\delta$ can be estimated using reweightings of the data. While biased splits are useful in the absence of multiple held-out samples, and have been proposed before \cite{karimi-etal-2015-squibs},\footnote{\newcite{karimi-etal-2015-squibs} discuss temporal  splits and splits based on neighbor-based heuristics that are similar in spirit to our worst-case splits.} they often overestimate performance in the wild. Our code is made publicly available at \url{https://github.com/google-research/google-research/tree/master/talk_about_random_splits}.

\section{Experiments}

We consider 7 different NLP tasks: POS tagging (like \newcite{Gorman:Bedrick:19}), two sentence representation probing tasks, headline generation, translation quality estimation, emoji prediction, and news classification. We experiment with these tasks, because they a) are diverse, b) have not been subject to decades of community-wide overfitting (with the exception of POS tagging), and c) three of them enabled temporal splits (see Appendix \S{A.5}).

\begin{table}\scriptsize \begin{tabular}{llll}
\toprule
{\bf Task}&{\bf Benchmark}&{\bf Source/Domain}&{\bf New Samples}\\
\midrule
{\sc POS Tagging}&WSJ&News&Xinhua\\
{\sc Probing}&SentEval&Toronto BC&Gutenberg\\
{\sc Emojis}&S140&Twitter&S140$^*$\\
{\sc QE}&WMT 2016&IT&WMT 2018\\
{\sc Headlines}&Gigaword&News&Gigaword$^*$\\
{\sc News}&UCI &News&UCI$^*$\\
\bottomrule
\end{tabular}\caption{Data used in our experiments. *: We time slice the original data to create different samples.}\label{tab:data}\end{table}

 \paragraph{Data splits} 

The datasets which we will use in our experiments are presented in Table~\ref{tab:data}. For all seven tasks, we will present results for standard splits when possible ({\sc POS}, {\sc Probing},{\sc QE}, {\sc Headlines}), random splits, heuristic and adversarial splits, as well as on new samples. In the case of {\sc Emojis}, {\sc Headlines} and {\sc News}, which are all time-stamped datasets, we leave out historically more recent data as our new samples. All {\bf new samples} are in-domain samples of data where models are supposed to generalize, i.e, samples from similar text sources.\footnote{Domains are commonly defined as collections of similar text sources \cite{Harbusch:ea:03,Koehn:Knowles:17}. In addition to using similar sources, we control for low $\mathcal{A}$-distance \cite{Ben-David:ea:06} by looking at separability; e.g., a simple linear classifier over frequent unigrams can distinguish between Penn Treebank development and test sections with an accuracy of 64\%; and between the development and our new sample with an accuracy of 69\%.}  This is a key point: These are {\em samples that any end user would expect decent NLP models to fair well on}. Examples include a sample of newspaper articles from newspaper $A$ for a POS tagger trained on articles from newspaper $B$; tweets sampled the day after the training data was sampled; or news headlines sampled from the same sources, but a year later. 

We resample {\bf random} splits multiple times (3-10 per task) and report average results. The {\bf heuristic} splits are obtained by finding a sentence length threshold and putting the long sentences in the test split. We choose a threshold so that approximately 10\%~of the data ends up in this split. The idea of checking whether models generalize to longer sentences is not new; on the contrary, this goes back, at least, to early formal studies of recurrent neural networks, e.g., \newcite{Siegelmann:Sontag:92}. In the \S{A.3}, we present a few experiments with alternative heuristic splits, but in our main experiments we limit ourselves to splits based on sentence length. 

Finally, the {\bf adversarial} splits are computed by approximately maximizing the Wasserstein distance between the splits. The Wasserstein distance is often used to measure divergence between distributions \cite{Arjovsky:ea:17,Tolstikhin:ea:18,Shen:ea:18,Shah:ea:18}, and while alternatives exist \cite{Ben-David:ea:06,BorGreRas+06a}, it is easy to compute and parameter-free. Since selecting {\em the} worst-case split is an NP-hard problem (e.g., by reduction of the knapsack problem), we have to rely on an approximation. We first compute a ball tree encoding the Wasserstein distances between the data points in our sample. We then {\em randomly}~select a centroid for our test split and find its $k$ nearest neighbors. Those $k$ nearest neighbors constitute our test split; %
the rest is used to train and validate our model. We repeat %
these steps to estimate performance on worst-case splits of our sample. See \S{A.4} for an algorithm sketch. Random, heuristic, and adversarial results are averaged across five runs.

\paragraph{POS tagging} We first consider the task in \newcite{Gorman:Bedrick:19}, experiment with heuristic and adversarial splits of the original Penn Treebank \cite{Marcus:ea:93}, and add the Xinhua section of OntoNotes 5.0\footnote{\url{https://catalog.ldc.upenn.edu/LDC2013T19}} as our {\bf New Sample}. Our tagger is NCRF$^{++}$ with default parameters.\footnote{\url{github.com/jiesutd/NCRFpp}} 

\paragraph{Probing} 
We also include two SentEval probing tasks \cite{Conneau:ea:18} with data from the Toronto Book Corpus: {\sc Probing-WC} (word classification) and {\sc Probing-BShift} (whether a bigram was swapped) \cite{Conneau:ea:18}. Unlike the other probing tasks, these two tasks do not rely on external syntactic parsers, which would otherwise introduce a new type of bias that we would have to take into account in our analysis. We use the official SentEval framework\footnote{\url{github.com/facebookresearch/SentEval}} and  
BERT \cite{Devlin:ea:19} as our sentence encoder. The probing model is a logistic regression classifier with $L_2$ regularization, 
tuned on the development set. 
As our {\bf New Samples}, we use five random samples of the 2018 Gutenberg Corpus\footnote{\url{tinyurl.com/zyq3yvn}} for each task, preprocessed in the same way as \newcite{Conneau:ea:18}.%

\renewcommand{\arraystretch}{1.2} %
\begin{table*}\small \centering \begin{tabular}{@{}llrrrrr@{}}
\toprule
&&\multicolumn{5}{c}{\bf Splits}\\
\cmidrule{3-7}
{\bf Task}&{\bf Model}&{\bf Standard}&{\bf Random}&{\bf Heuristic}&{\bf Adversarial}&{\bf New Samples}\\
\midrule
{\sc POS Tagging}&{\sc NCRF$^{++}$}&0.961&0.962&0.960&0.944&{\bf 0.927}\\
\midrule
{\sc Probing-WC}
&\multirow{2}{*}{BERT}&0.520&0.527&{\bf 0.232}&0.250&0.279\\

{\sc Probing-BShift}
&&0.800&0.808&0.695&0.706&{\bf 0.450}\\
\midrule
{\sc Headline Generation}$^*$%
&seq2seq& 0.073 & 0.095 & 0.062 & {\bf0.040} & 0.069 \\
\midrule
{\sc Quality Estimation}$^\dagger$
&\multirow{3}{*}{MLP-Laser}&0.502&0.626&0.621&0.711&{\bf 0.767}\\
{\sc Emoji Prediction}
&&-&0.125&0.196&{\bf -0.040}&0.091\\

{\sc News Classification}&&-&0.681&0.720&0.634&{\bf 0.618}\\
\midrule %
MSE ({\bf New Samples})&&0.179&0.030&0.015&0.011&-\\
\bottomrule
\end{tabular}
\caption{Error reductions over random baselines on {\bf Standard} (original) splits, if available, {\bf Random} splits (obtained using cross-validation), {\bf Heuristic} splits resulting from a sentence length-based threshold, {\bf Adversarial} splits based on (five) approximate maximizations of Wasserstein differences between splits, and on {\bf New Samples}. We bold face the {\bf lowest} error reduction, i.e., where results differ the most from the random baseline. {\bf We see that standard and random splits consistently over-estimate {\em real} performance on new samples, which is sometimes even lower than performance on adversarial splits.} We also report the mean squared error (MSE) with respect to {\bf New Samples}, which shows {\bf Adversarial} estimates empirical error best. Note: {\em While annotator bias could explain POS tagging results, there is no annotator bias in the other tasks.} $^*$: For {\sc Headlines} we use an identity baseline. Scores are ROUGE-2; see \S{A.1} for more. $^\dagger$: For {\sc Quality Estimation}, we report RMSE. The WMT QE 2014 best system obtained RMSE of 0.64; our system is significantly better with 0.50 on the standard split. %
}\label{tab:results}\end{table*}

\paragraph{Quality estimation}
We use the WMT 2014 shared task datasets for {\sc Quality Estimation}. Specifically, we use the Spanish-English data from Task 1.1: scoring for perceived post-editing effort. The dataset comes with a training and test set, and a second, unofficial test set, which we use as our {\bf New Sample}. In the \S{A.2}, we also present results training on Spanish-English and evaluating on German-English. We present a simple model that only considers the target sentence, but performs better than the best shared task systems: we train an MLP over a LASER sentence embedding \cite{Schwenk:ea:19} with the following hyper-parameters: two hidden layers with 100 parameters each and ReLU activation functions, trained using the Adam stochastic gradient-based optimizer \cite{Kingma:Ba:15}, a batch size of 200, and $L_2$ penalty of strength $\alpha=0.01$.

\paragraph{Headline generation} We use the standard dataset for headline generation, derived from the Gigaword corpus \cite{napoles12}, as published by \newcite{rush15}. The task is to generate a headline from the first sentence of a news article. Our architecture is a sequence-to-sequence model with stacked bi-directional LSTMs with dropout, attention \cite{luong15} and beam decoding; the number of hidden units is 128; we do not pre-train. Different from \newcite{rush15}, we use subword units \cite{sennrich16} to overcome the OOV problem and speed up training. The ROUGE scores we obtain on the standard splits are higher than those reported by \newcite{rush15} and comparable to those of \newcite{nallapati16}, e.g., ROUGE-1 of 0.321. 
As our \textbf{New sample}, we reserve 20,000 sentence-headline pairs each from the first and second halves of 2004 for validation and testing; years 1998-2003 are used for training. 
For all the experiments we report the error reduction in ROUGE-2 of the model over the \textsc{identity} baseline, which simply copies the input sentence (other ROUGE values are reported in the \S{A.1}). In \S5, we will explore how much of a performance drop on the fixed test set is caused by shifting the \emph{training} data by only five years to the past.

\paragraph{Emoji prediction} \newcite{Go:ea:09} introduce an emoji prediction dataset, collected from Twitter and is time-stamped. 
We use 
the 67,980 tweets from June 16 as our {\bf New Sample}, and tweets from all previous days for the remaining experiments. 
For this task, we again train an MLP over a LASER embedding \cite{Schwenk:ea:19} with hyper-parameters: 
two hidden layers with 50 parameters each and ReLU activation functions, trained using the Adam stochastic gradient-based optimizer \cite{Kingma:Ba:15}, a batch size of 200, and $L_2$ penalty of strength $\alpha=0.01$. See \S5 for a discussion of temporal drift in this data.

\paragraph{News classification} We use a UCI Machine Learning Repository text classification problem.\footnote{\url{tinyurl.com/yysysmtr}}  Our datapoints are headlines associated with five different news genres. We use the last year of this corpus as our {\bf New Sample}. We sample 100,000 headlines from the rest and train an MLP over a LASER embedding \cite{Schwenk:ea:19} with the following hyper-parameters: two hidden layers with 100 parameters and ReLU activation functions, trained using the Adam stochastic gradient-based optimizer \cite{Kingma:Ba:15}, dynamic batch sizes, and $L_2$ penalty of strength $\alpha=0.01$. 

\section{Results}

Our results are presented in Table~\ref{tab:results}.
Since the results are computed on different subsamples of data, we report {\em error reductions over multinomial random} (or, for {\sc Headlines}, {\em identity}) {\em baselines}, following previous work comparing system rankings across different samples \cite{Soegaard:13}.
More formally, we present error reduction as $r=\frac{p_s-p_b}{1-p_b}$%
, where $p_{s}$ and $p_{b}$ are the performances of the system at hand and the multinomial random baseline. %

Our main observations are the following: (a) \textbf{Random splits (and standard splits) consistently under-estimate error on new samples}. The absolute differences between error reductions over random baselines for random splits and on new samples are often higher than 20\%, and in the case of {\sc Probing-BShift}, for example, the BERT model reduces 80\% of the error of a random baseline when data is randomly split, but only 45\%~averaging over five samples of new data from the same domain. (b) Heuristic splits sometimes under-estimate error on new samples. Our heuristic splits in the above experiments are quite aggressive. We only evaluate our models on sentences that are longer than any of the sentences observed during training. Nevertheless for {5/7~tasks}, this leads to more optimistic performance estimates than evaluating on {\em new}~samples! 
(c) The same story holds for adversarial splits based on approximate maximization of Wasserstein distances between training and test data. While adversarial splits are very challenging, results on adversarial splits are more optimistic than on new samples in {4/7 cases}. Note the fact that random splits over-estimate real-life performance also leads to misleading {\bf system rankings}. If, for example, we remove the CRF inference layer from our POS tagger, performance on our {\bf Random} splits drops to 0.952; on the {\bf New Sample}, however, performance is 0.930, which is significantly better than {\em with} a CRF layer.%

\paragraph{Discussion} In the spirit of earlier work \cite{Sakaguchi:ea:17,Madnani:Cahill:18,Gorman:Bedrick:19}, we provide recommendations for future evaluation protocols: %
(i) In the absence of multiple held-out samples, using \textbf{biased splits} better approximates real-world performance and can help determine what data characteristics affect performance. %
(ii) Evaluating on %
\textbf{new samples} %
is superior and also enables significance testing across datasets \cite{Demsar:06}, providing confidence estimates. %
Several benchmarks already provide multiple, diverse test sets \citep[e.g.][]{hovy-etal-2006-ontonotes,parsingweb2012sharedtask,williams-etal-2018-broad}; we hope more will follow. %
What explains the high variance across samples in NLP? One reason is the dimensionality of language \cite{Bengio:ea:03}, but in \S{A.5} we also show significant impact of temporal drift. 

\paragraph{Conclusions} 
We have shown that out-of-sample error can be hard to estimate from random splits, which tend to underestimate error by some margin, but even biased and adversarial splits sometimes underestimate error on new samples. We show this phenomenon across seven very different NLP tasks and provide practical recommendations on how to best bridge the gap between experimental practices and what is needed to produce truly robust NLP models that perform well in the wild. 

\section*{Acknowledgments}

We would like to thank our reviewers for their comments and for engaging in an interesting discussion. The paper also benefited greatly from discussions with several of our colleagues at Google Research, including Slav Petrov and Sascha Rothe. 

\bibliography{ms}
\bibliographystyle{acl_natbib}

\newpage
\appendix

\section{Appendices}
\label{sec:appendix}

We present supplementary details about two of our tasks in \S{A.1} and \S{A.2} and discuss variations over heuristic splits in \S{A.3}. In \S{A.4}, we present the pseudo-algorithm for how we compute adversarial splits, and finally, in \S{A.5}, we present our results documenting temporal drift. 

\subsection{Headlines}

\begin{figure}\centering 
  \includegraphics[width=\columnwidth]{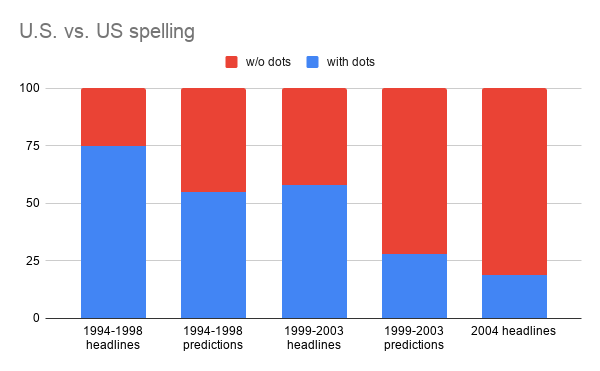}
\caption{Proportions of \textit{US} vs.\ \textit{U.S.} spellings in the headlines for two training sets and the test set (2004) as well as in the two models' predictions on the test set.} \label{fig:app:headline}\end{figure}

Table \ref{tab:app:headline} reports the error reduction in ROUGE-1, ROUGE-2 and ROUGE-L over the identity baseline (see \S2) for the different data splits. The results are consistent with Table \ref{tab:results}. Figure \ref{fig:app:headline} gives more details on an interesting drift phenomenon, which contributed to the superior performance of the model trained on the most recent five years (1999-2003). Apparently, the dotless spelling of {\em U.S./US} ('United States') became more common over time. %
Consequently, the model trained on the 1999-2003 part generated \textit{US} more frequently than the model trained on 1994-1998.

\begin{table}\centering \scriptsize \begin{tabular}{lrrr}
\toprule
& {\sc rouge-1} & {\sc rouge-2} & {\sc rouge-l} \\
\midrule
{\bf Standard}& 0.080 & 0.073 & 0.097  \\
\midrule
{\bf Random}& 0.109 & 0.095 & 0.127 \\
{\bf Heuristic}       & 0.091 & 0.062 & 0.109 \\
{\bf Adversarial}        & 0.060 & 0.040 & 0.080 \\
{\bf New Sample}& 0.067 & 0.069 & 0.091\\
\bottomrule
\end{tabular}\caption{Error reduction as compared with an identity baseline (output as input) for three ROUGE metrics. Random is a five-fold cross-validation result.} \label{tab:app:headline}\end{table}

\subsection{Quality Estimation}

In the results above, we train and test our quality estimation regressor on Spanish-English from WMT QE 2014. We also ran a similar experiment where we used the German-English test data as our {\bf New Sample}. Here, we see a similar pattern to the one above: The RMSE on the {\bf Standard} split was 0.630, which is slightly higher than for Spanish-English; with our {\bf Heuristic} split, RMSE is 0.652; for {\bf Adversarial}, it is 0.626 (which is slightly better than with standard splits), and on our {\bf New Sample}, RMSE is 0.813.  

\begin{table*}\centering \begin{tabular}{@{}llrrrr@{}}
\toprule
&&\multicolumn{4}{c}{\bf Splits}\\
\cmidrule{3-6}
{\bf Task}&{\bf Model}&{\bf Standard}&{\bf Bootstrap}&{\bf Random Length}&{\bf Rare Words}\\
\midrule
{\sc Probing-WC}%
&\multirow{2}{*}{BERT}&0.520&0.504&0.554&0.339\\
{\sc Probing-BShift}%
&&0.800&0.807&0.798&0.731\\
\bottomrule
\end{tabular}
\caption{Error reductions over random baselines on {\bf Standard} (original) splits, if available, {\bf Bootstrap} splits, {\bf Random Length} splits resulting from a sentence length-based separation, {\bf Rare Words} splits based on word frequency.
}\label{tab:alternative_heuristics}\end{table*}

\subsection{Alternative Heuristic Splits}
For both SentEval tasks we experimented with the following alternatives for heuristic splits.

\paragraph{Bootstrap Resampling} Instead of cross-validation, a random split can be generated by bootstrap resampling. For this we randomly select 10\% of the data as test set and then randomly sample (with replacement) a new training and dev set from the remaining examples.

\paragraph{Random Length} As alternative to the length threshold heuristic in earlier experiments we randomly sample a length and select all examples having this length to be part of the test set. We repeat this procedure until approximately 10\% of the data ends up in the test set. With this procedure we create 5 different test sets. We included this heuristic in order to see how fragile the probing setup is.

\paragraph{Rare Words} Another alternative for heuristic splits is to use word frequency information. Here we assign those sentences containing at least one of the rarest words of the dataset to the test set. This way we end up again with approximately 10\% of the data in the test set. Note that this way we create only 1 dataset, because it's not a random process.

\paragraph{Results} Table~\ref{tab:alternative_heuristics} lists the results. While bootstrap resampling leads to slightly lower error reduction than cross-validation we decided to report the latter in the main part of this paper, because it is a more wide-spread way to randomly split datasets. Random Length results are comparable to standard splits results. The split based on word frequency (Rare Words) leads to considerable drop in both tasks. However, it is not as strong as the drop of the heuristic split (length threshold) in the main part of the paper.

\begin{figure}
    \centering
    \includegraphics[width=\columnwidth]{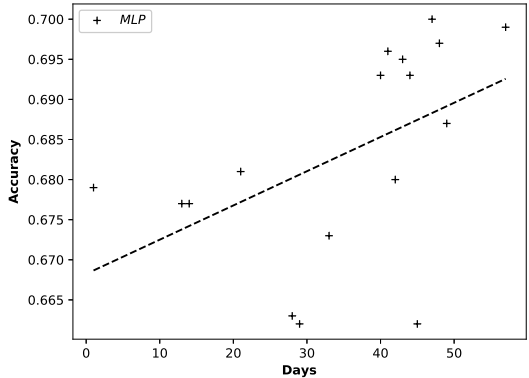}
    
    \caption{Temporal drift in emoji prediction. The correlation between temporal gap and performance is significant ($p<0.05$).}
    \label{fig:presidents}
\end{figure}

\begin{algorithm}
 \KwData{Dataset $G_{\mathit{train}}$}
 \KwResult{Adversarial split: $G_{\mathit{train}}$ and $G_{\mathit{test}}$}
 $T\leftarrow \mathit{WassersteinBallTree}(G_{\mathit{train}})$\;
 $G_{\mathit{test}}\leftarrow \{\langle x,y\rangle\sim G_{\mathit{train}}\}$\;
 $G_{\mathit{train}}\leftarrow G_{\mathit{train}}/\{\langle x,y\rangle\}$\;
 \While{$i\leq k$}{
 $\langle x_i,y_i\rangle = \min_{\langle x',y'\rangle}T(x,x')$\;
 $G_{\mathit{train}}\leftarrow G_{\mathit{train}}/\{\langle x_i,y_i\rangle\}$\;
 $G_{\mathit{test}}\leftarrow G_{\mathit{test}}\cup  \{\langle x_i,y_i\rangle\}$\;
 $i+=1$\;
   }
 \caption{\label{alg} Computing adversarial splits}
\end{algorithm}

\subsection{Computing adversarial splits}

We present the pseudo-algorithm of our implementation of approximate Wasserstein splitting in Algorithm~\ref{alg}. We also make the corresponding code available as part of our code repository for this paper.

\subsection{The significance of drift}

Some of our splits in the main experiments were based on slicing data into different time periods ({\sc Headlines}, {\sc Emojis}). Since temporal drift is a potential explanation for sampling bias, we analyze this in more detail here. We show that temporal drift is pervasive and leads to surprising drops in performance. We note, however, that temporal drift is not the only cause of sampling bias, of course. Since we have time stamps for two of our datasets we study these in greater detail. For similar studies of temporal drift, see \citet{lukes-sogaard-2018-sentiment,rijhwani-preotiuc-pietro-2020-temporally}.

\paragraph{Headline generation} Our headline generation data covers the years 1994 to 2004. Having reserved 20,000 sentence-headline pairs from the first half of 2004 for validation and the same amount from the second half for testing, we use 50\% of the years 1994-2003 for training three models. The models' architectures and parameters are identical (same as in Sec.\ 3). The only difference is in what the models are trained on: (a) a random half, (b) the first, or (c) the second half of 1994-2003. The training data sizes are comparable (1.63-1.76M), the publisher distributions (AFP, APW, CNA, NYT or XIN) are also similar. Hence, the models are expected to perform similarly on the same test set. 

\begin{table}\centering \scriptsize \begin{tabular}{lrrr}
\toprule
& {\sc rouge-1} & {\sc rouge-2} & {\sc rouge-l} \\
\midrule
{\bf Identity baseline}& 0.302 & 0.110 & 0.260 \\
\midrule
{\bf 50\% of 1994-2003}& 0.409 & 0.205 & 0.386 \\
{\bf 1994-1998 }       & 0.346 & 0.161 & 0.329 \\
{\bf 1999-2003}        & 0.413 & 0.208 & 0.388 \\
\bottomrule
\end{tabular}\caption{Performance of three seq2seq models trained on different samples, evaluated on 2004 data.}\label{tab:headline}\end{table}

As Table~\ref{tab:headline} indicates, shifting the training data by five years to the past results in a big performance drop. Sampling training data randomly or taking the most recent period produces models with similar ROUGE scores, both much better than the identity baseline. However, about half of the gap to the identity baseline disappears when older training data is taken. %
In the \S{A.1}, we give an example of temporal drift in the {\sc Headlines} data: \textit{US} largely replaces \textit{U.S.} in the newer training set and the test set.

\paragraph{Emoji prediction} For emoji prediction, \newcite{Go:ea:09} provide data for a temporal span of 62 days. We split the data into single days and keep the splits with more than 25,000 datapoints in which both classes are represented. We use the last of these, June 16, as our test sample and vary the training data from the first day to the day before June 16. Figure~\ref{fig:presidents} (left) visualizes the results.

\end{document}